\documentclass[conference]{IEEEtran}
\IEEEoverridecommandlockouts

\usepackage{cite}
\usepackage{amsmath,amssymb,amsfonts}
\usepackage[ruled,vlined]{algorithm2e}
\usepackage{graphicx}
\usepackage{textcomp}
\usepackage{xcolor}
\usepackage{tabularx}
\usepackage{geometry}

\def\BibTeX{{\rm B\kern-.05em{\sc i\kern-.025em b}\kern-.08em
    T\kern-.1667em\lower.7ex\hbox{E}\kern-.125emX}}
\begin{document}

\title{A Generalized Kalman Filter Augmented Deep-Learning based Approach for Autonomous Landing in MAVs \\

\thanks{NG acknowledges an NVIDIA GPU grant and Associateship sponsored by the International Center for Theoretical Physics \newline
 978-1-6654-3323-5/21/\$31.00 ©2021 IEEE
}
}

\newgeometry{top=25.4mm, bottom=19mm,left=19mm,right=19mm} 
\author{\IEEEauthorblockN{Pranay Mathur$^1$}
\IEEEauthorblockA{\textit{Dept. of EEE} \\
\textit{BITS Pilani, K.K Birla Goa Campus}\\
Goa, India \\
f20170487@goa.bits-pilani.ac.in}
\and
\IEEEauthorblockN{Yash Jangir$^2$}
\IEEEauthorblockA{\textit{Dept. of EEE} \\
\textit{BITS Pilani, K.K Birla Goa Campus}\\
Goa, India  \\
f20190526$@$goa.bits-pilani.ac.in}
\and
\IEEEauthorblockN{Neena Goveas$^3$}
\IEEEauthorblockA{\textit{Department of Computer Science} \\
\textit{BITS Pilani, K.K Birla Goa Campus}\\
Goa, India \\
neena@goa.bits-pilani.ac.in}
}

\maketitle

\begin{abstract}
Autonomous landing systems for Micro Aerial Vehicle (MAV) have been proposed using various combinations of GPS based, vision and fiducial tag-based schemes. Landing is a critical activity that a MAV performs and poor resolution of GPS, degraded camera images, fiducial tags not meeting required specifications and environmental factors pose challenges. An ideal solution to MAV landing should account for these challenges and for operational challenges which could cause unplanned movements and landings. Most approaches do not attempt to solve this general problem but look at restricted sub-problems with at least one well-defined parameter. 
In this work, we propose a generalized end-to-end landing site detection system using a two-stage training mechanism, which makes no pre-assumption about the landing site. Experimental results show that we achieve comparable accuracy and outperform existing methods for the time required for landing.  
\end{abstract}


\section{Introduction}
Recent years have seen the rapid development and maturity of Micro Aerial Vehicle (MAV) technology as they are agile, easily maneuverable and offer cost effective solutions. The constraints of cost and weight in MAV has meant that many of the techniques developed for larger Unmanned Aerial Vehicles (UAV) cannot be used as is. One of the main concerns in MAV research is to design safe landing techniques in unfavorable situations where GPS based precise localization is not feasible. Vision based techniques which are commonly used, have to work within the constraints of a non ideal camera and image capture and account for partial or complete occlusion of the target site during the landing operation. 

In this work, we propose a novel two-stage trained deep learning-based method that achieves accurate vision-based landing. There are no assumptions made regarding the landing site or any symbols placed on it. A short training time ensures our model can also be be used in situations where the image of the landing site are available just prior to flight. We have successfully tested our solution during landing operations where there was occlusion and image distortion. 

We find that our proposed technique gives comparable performance in terms of accuracy and outperforms other solutions in terms of the time required for landing. The contributions of this paper are as follows:
\begin{itemize}
  \item An implementation and training strategy for a deep-learning framework that allows passing images of an arbitrary landing site prior to flight and makes no assumptions on the contours, profile or texture of this site.
  \item A customized Kalman Filter based controller that utilises the constraints imposed by the bounding box coordinates to achieve precise control during landing.
\end{itemize}
In addition, we have also created a custom dataset for training our network initially prior to using augmented dataset of the actual landing site images for the second stage of training. 

\section{Related Works}\label{RW}
UAVs have found their way into various applications involving military, delivery, search and rescue operations. Six-degrees of freedom and their agile behaviour make them an ideal tool for critical operations in inaccessible regions. For many missions, certain scenarios lead to further constraints in the form of lack of infrastructure, requiring smaller and cheaper devices\cite{b1}. 
MAVs are cost-effective and agile but lack of on-board computational resources and peripherals, results in the need for customized software which can ensure smooth functioning. One of the basic requirements for a MAV is autonomous landing in a precise manner without depending on external support, for example, a positioning method based on the global positioning system (GPS)\cite{ba} or pre-decided image tags to be kept on the landing site. Autonomous landing using images captured by on-board cameras is one of the options which has provided promising results\cite{bc}.

There exist a host of previous research work attempting to tackle the issue of vision-guided autonomous MAV landing. A broad categorization can be performed on the basis of \newgeometry{top=19mm, bottom=19mm,left=19mm,right=19mm}
whether the visual sensor used is on-board\cite{b3,b4,b5} or off-board\cite{b6,b7,b8}. We focus on on-board visual system approaches, which are of two kinds, classical image processing techniques such as contour recognition and corner detection \cite{c3,c5,c6} or learning based techniques \cite{c1,c2}.

Autonomous vision based landing for MAVs consists of two phases, first involves identification of landing sites and the second involves visual estimation containing position estimation and direction estimation to generate a path towards the landing site. Issues affecting the detection of the landing site include the fact that the captured image of the landing site is subject to affine transformation, scale change, rotation, translation and oblique projection transformation\cite{bb}. Moreover environmental factors such as vibration and uneven illumination give rise to degradation of the quality of images captured. Most techniques make strong assumptions regarding the illumination of the environment and the contours and profile of the tags used. An ideal approach should perform well in the above non-ideal circumstances and cases where occlusion of the landing site occurs. 

The issues related to direction estimation and correction process during the landing operation is typically controlled by a Kalman fiter based approach. Existing methods that use Kalman filters with inputs from classical vision based techniques have multiple drawbacks, such as the tracking error accumulating \cite{kalm}.  
Most proposed techniques make assumptions that all details of the landing site are known well in advance and the MAV will not encounter any unknown parameter. This results in loss of flexibility and imposing infrastructural requirements on the end-user.

In this work we focus on on-board vision based solutions for an arbitrary landing site with no assumptions regarding image capturing system. Our approach is able to handle scenarios where partial occlusion of the landing site occurs and the landing vehicle field of view includes only partial views of the landing site in the captured images.

\section{Methodology}
In this study, an end-to-end deep-learning strategy is proposed to enable on-board vision based landing for arbitrary landing sites. Our approach consists of the  following - the first phase where we deploy a target bounding box detection network to extract coordinates of the landing site in an image captured by the MAV. This is explained in Section \ref{LSD}. The deep-learning based method has a two-stage training strategy which is detailed in Section \ref{train}. Our method is designed for high-speed and accurate detection of arbitrary landing sites. 
In the second phase, we use these detected coordinates and a customized Kalman Filter based controller to perform landing. This is detailed in Section \ref{EKF}. 

\subsection{Landing Site Detection}\label{LSD}
Identification of the landing site in images captured by the MAV is a critical step for performing autonomous landing. We pre-process our images by conversion to 8-bit grayscale images and use CLAHE\cite{b0}, an equalization technique that tackles the problem of over-amplification of the contrast and makes our system more robust to changes in environmental illumination. Instead of using a pre-defined fiducial tag or set of known landing sites we use an arbitrary image of the landing site. This is enabled by our model architecture detailed in Section \ref{MA} and our customized training strategy in Section \ref{train}.   

\subsubsection{Model Architecture}\label{MA}
To achieve the goal of fast detection of an arbitrary image of the landing site available only prior to the flight, we are constrained in the total time available for training of our network. We solve this by pre-training our network and utilizing transfer learning.

We employ the Faster R-CNN\cite{frcnn} model architecture for accurate and fast estimation of the bounding-box coordinates. It comprises of two modules, a deep fully convolutional network that proposes regions, referred to as the Region Proposal Networks(RPN), and a Fast R-CNN\cite{fast} detector that extracts features from the proposed regions. This technique is appropriate for our case in the context of Transfer Learning where we are constrained to train only on a small dataset, in our particular case the actual landing site images, using the weights of a network trained on a larger sized dataset, our custom dataset, detailed in \ref{DS}.
The input images from the MAV are represented as $Height \times Width \times Depth$ tensors, which are passed through a pipeline consisting of a pre-trained Convolutional Neural Network layers and end with a convolutional feature map. This is used as a feature extractor for the RPN. The Region Proposal Network uses the features that the CNN computes to find probable landing sites in the image. This number can be pre-defined with an upper bound. The RPN uses fixed sized reference bounding boxes, referred to as anchors, placed uniformly throughout the original images. These are the regions checked for containing relevant objects and their location is noted. The features extracted by the CNN are used and Region of Interest (RoI) Pooling is applied to relevant locations. 
The final module is the R-CNN module, which uses the previous information to classify the content in the bounding box and adjust the bounding box coordinates. Finally a post-processing step to prevent duplicate proposals is applied. This algorithmic approach is referred to as Non-Maximum Suppression (NMS). These bounding-box coordinates are then passed to our controller.

We find that our choice of the network architecture gives superior performance in-terms of accuracy after transfer learning, size of required dataset, required time for training, and time for inference.

\subsection{Training}\label{train}
Training is an essential step to tune the weights of a model framework for accurate inference but poses a major challenge when the landing site images are received just prior to flight. This lack of time imposes constraints on number of epochs that can be used for training. Additionally, we are constrained to perform this training on the on-board computer and not on a computationally powerful server. 

We propose a novel two stage strategy to solve this problem.
In the first stage we train our network on our custom dataset detailed in \ref{DS}. This dataset contains a wide variety of possible landing sites in a variety of non-ideal scenarios and environmental conditions and is split in a 70:20:10 ratio of training, validation and test data. 
Initially, we train the network on our custom dataset till we achieve a $loss$ tending to $unity$. This is done to adjust and provide an accurate initial estimate of the graph weights. This coarse tuning is performed in order to achieve a reduction in the time required for training on the actual site without compromising on accuracy.
The second stage involves finer adjustment of the weights by training our model pre-flight and is performed using images of the actual site. Image augmentation is used to expand the size of the dataset and improve performance. This involves horizontal and vertical shift, random rotation, random zoom and random brightness augmentation. Training is performed on this dataset using the weights of the graph obtained from the previous step till the $loss$ tends to 0.5. We find that this heuristically determined value is a limit below which we get accurate results. This stage presents a trade-off between training times and resultant accuracy. These have been discussed in \ref{eval}.

\subsection{Kalman Filter based Controller}\label{EKF}
\begin{figure}[h]
\centerline{\includegraphics[width=0.45\textwidth]{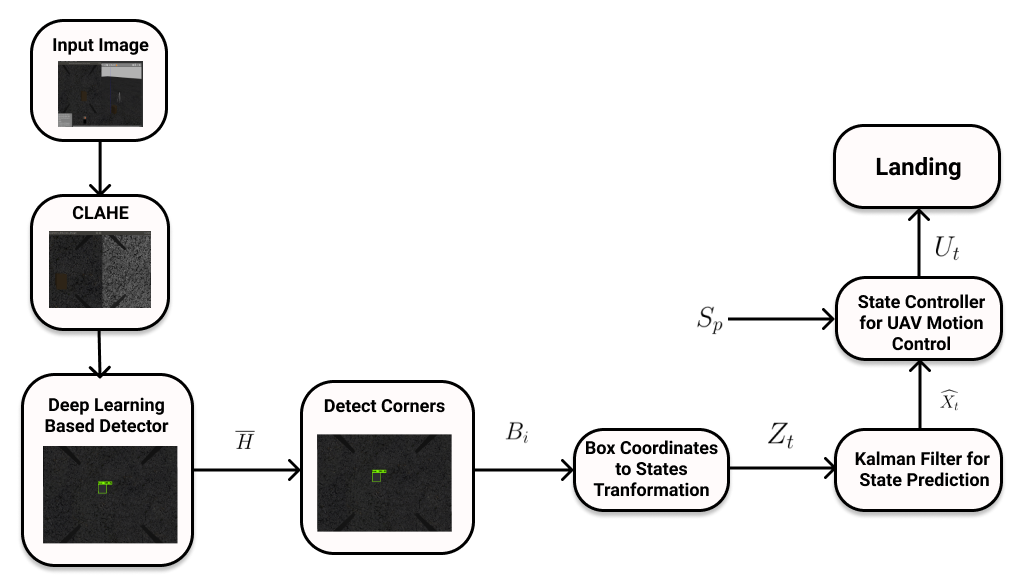}}
\caption{A block diagram of working}
\label{bd}
\end{figure}

We use a customized Kalman filter based controller built upon the work of \cite{EKF_git} for precisely maneuvering the quadcopter to the landing site. In our method, instead of using fiducial tags for landing site detection\cite{EKF_git}, we use Kalman filter based controller with bounding-box coordinates from our model architecture. A bounding-box gives comparatively less information about pose of the landing site than is received from fiducial tags. Additionally, in case of partially visible or occluded landing sites, fiducial tag based methods cannot work as the tag must be fully visible for detection. In contrast, our controller will use the best available coordinates and thus will work even in case of partially occluded images.\\
Our method consists of passing the bounding box coordinates of the landing site detected and transforming bounding-box coordinates to states, to act as an input to the kalman filter. Resulting predicted states are passed to state controller which calculates errors and generates subsequent motion control commands for the MAV as shown in Fig.\ref{bd}. 

\newenvironment{conditions*}
  {\par\vspace{\abovedisplayskip}\noindent
   \tabularx{\columnwidth}{>{$}l<{$} @{${}:{}$} >{\raggedright\arraybackslash}X}}
  {\endtabularx\par\vspace{\belowdisplayskip}}

In Fig.3 :
\begin{conditions*}
\overline{H} & Detected Observations by Deep learning Model\\
B_{t} & Detected Corners\\
Z_{t} & Transformed states \\
\widehat{X_{t}} & Predicted states by Kalman Filter\\
U_{t} & Generated motion control signals by state controller\\
\end{conditions*}

\subsubsection{Transformation of Bounding Box Coordinates to States}
We receive the bounding box coordinates of the target landing site from the first stage. We define these bounding box coordinates as $(X_{i},Y_{i})$ for $i \in \{0,1,2,3\} $.

These coordinates are used to calculate the states, $Z_{t}$, which will be passed to the Kalman Filter. These states consist of the centroid coordinates of bounding box, $X_{c}$ and $Y_{c}$, the angle between the landing site and horizontal of image, $\theta$ and the width and height of the landing site $W_s$ \& $H_s$ respectively. For the Angle between the landing site and horizontal of image ($\theta$), we calculate Vectors : 
\begin{equation}
\begin{split}
 \overrightarrow{X} = \overrightarrow{X_{1}} - \overrightarrow{X_{0}}\\
 \overrightarrow{Y} = \overrightarrow{Y_{1}} - \overrightarrow{Y_{0}}
\end{split}
\end{equation}
where ($X_{0},Y_{0}$) and ($X_{1},Y_{1}$) are bounding box coordinates $(X_{i},Y_{i})$ for $i \in \{0,1\} $ 

Calculation of the angle between these two vectors gives us the value of $\theta$.\\
 If $E_{P_{1},P_{2}}$ is euclidean distance between any two points in 2-D Cartesian coordinate system then $W$ is calculated by averaging two parallel euclidean widths of box and $H$ calculated by averaging two parallel euclidean heights of box.

\begin{equation}
\begin{split}
   W = \dfrac{E_{B_{0},B_{1}} + E_{B_{2},B_{3}} }{2}\\
   H = \dfrac{E_{B_{0},B_{2}} + E_{B_{1},B_{3}} }{2} 
\end{split}
\end{equation}

\subsubsection{State Controller}
We use a Proportional Integral Derivative Controller(PID)\cite{K_PID} to navigate to the target states with feedback from the predicted states by the Kalman Filter. Predicted States corresponding target landing site in the image are received periodically from Kalman Filter.
Unlike aruco tags or any other fiducial tags where a complete view of the tag would be required, in our method even in cases where the landing site is partially visible, the Kalman Filter uses the available coordinates to accomplish a successful landing.

To control of X,Y and $\theta$, We define centroid coordinates of image frame as :
\begin{conditions*}
X_{i_c} & X Coordinate of the centroid of image frame\\
Y_{i_c} & Y Coordinate of the centroid of image frame\\
\end{conditions*}

Closed loop PID controllers are used to control the position along X and Y axis and the orientation with respect to landing site. In order to achieve the target landing site position,the image frame centroid, $X_{i_c},Y_{i_c}$ must coincide with the landing site centroid prediction by Kalman Filter. PID controllers generate appropriate velocity commands for the Autopilot using the landing site centroid coordinates for feedback. \par

In aruco tags and other fiducial tag based methods extra information in the form of pose of the tag which includes orientation is received, however this information is absent in bounding box coordinates.\\
In our method for orientation, Angle between the landing site and horizontal of image, $\theta$  which is calculated by bounding box coordinates, is used. Since we use a bounding box which has a symmetric orientation on rotation of $90^{o}$, the closest direction parallel to the edges is considered 0. $\theta$ is directly the error for PID controller. In order to reach the correct orientation $\theta$ must tend to zero. Hence, PID controller generates control signals as yaw rate commands for the MAV to achieve correct orientation with respect to the target landing site.\\

\begin{figure}[h]
\centerline{\includegraphics[width=0.45\textwidth]{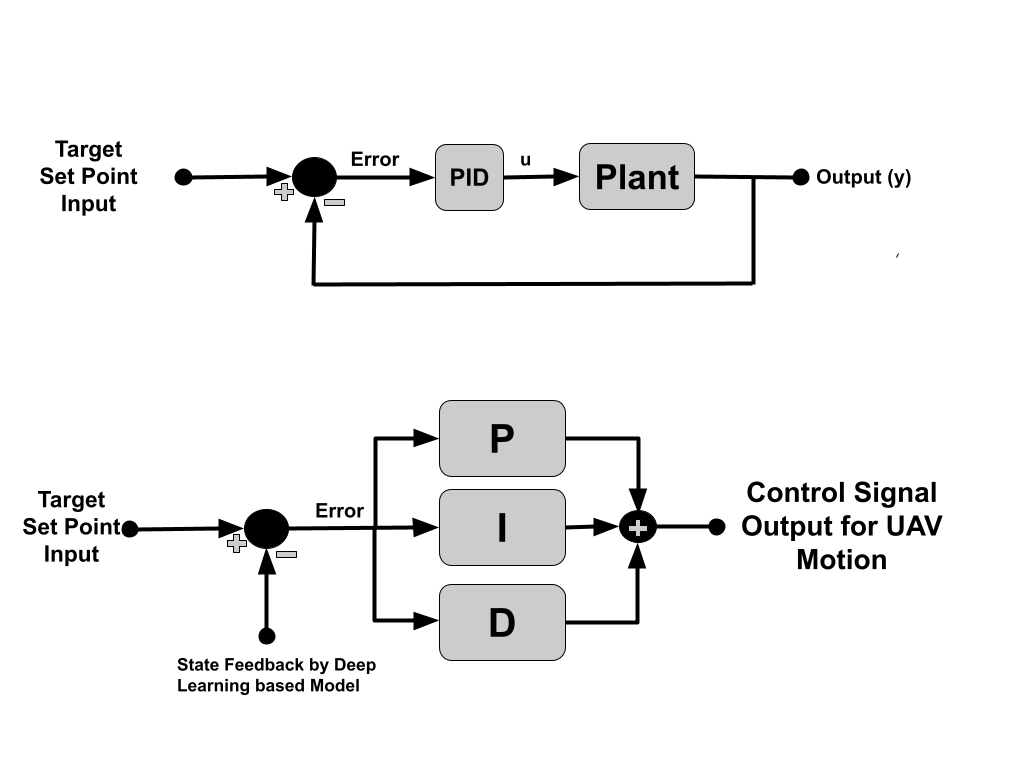}}
\caption{ Closed Loop PID controller}
\label{fig}
\end{figure}

Control of altitude $Z$: While the position along X and Y axis and the orientation are controlled by PID controllers, the altitude cannot directly be controlled by a PID controller. 
As the MAV descends down, the width, $W$ and height, $H$ of the target landing site increases as going down will increase the size of object as seen in camera image. This shows a dependency of altitude on $W$ and $H$. So, we estimate this dependency in altitude as: 
\begin{equation}
\begin{split}
 Z_{e} = K*|W - H|
\end{split}
\end{equation}
where $K$ represents the proportionality constant.

We define a condition based on error in altitude,$Z_{e}$ and current altitude of MAV, $Z_{i}$. While this condition is satisfied, the MAV continues descending by a predefined altitude, $Z_{f}$. We generate new target altitude position, $Z_{pos}$ for the MAV:
\begin{equation}
\begin{split}
Z_{pos} = Z_{i} - Z_{f}
\end{split}
\end{equation}
 
\par Hence, we are able to generate motion control commands with the predicted states of the target landing site, the MAV aligns itself with landing site and descends down using the State controller to reach to the landing site.

\subsubsection{Final Landing Sequence}
In order to land safely on the target site, the MAV must be in the correct position, orientation and altitude, with respect to the target landing site. At the instant the pose aligns with the target landing site and the altitude is below a pre-defined threshold of 0.3 m, landing sequence is initiated by the State controller. The state controller terminates publishing control commands and descent at a controlled rate in a linear trajectory is carried out.

\section{Implementation}
We implement and evaluate our proposed solution in the Gazebo Simulation environment using the Robot Operating System (ROS) framework\cite{ros}. The infrastructure and hardware simulated is ubiquitously available and our implementation is agnostic to hardware specification. 

\subsection{Dataset}
\label{DS}
We use a custom built dataset to train our model on. These are obtained primarily from simulations carried out in gazebo. The images are collected across a variety of environmental conditions such as excessive lighting, differential lighting, partial occluded sites and across different landing sites including simple boxes and white pads against backgrounds such as grass patches, roads and dirt patches. The dataset contains 450 images. A representative subset of the images collected to counter the problem of merging backgrounds with minute color and contrast difference have been shown in Fig. \ref{dataset}. We highlight the fact that we do not intend to train the model solely for this particular dataset, merely alter and give an initial estimate of the weights in the graph so training may be faster on images of the actual landing site. An accurate initial estimate of weights are achieved using comparatively fewer number of images when compared to a conventional dataset. 

\subsection{Simulation Environment}
\begin{figure}[h]
\centerline{\includegraphics[width=0.45\textwidth]{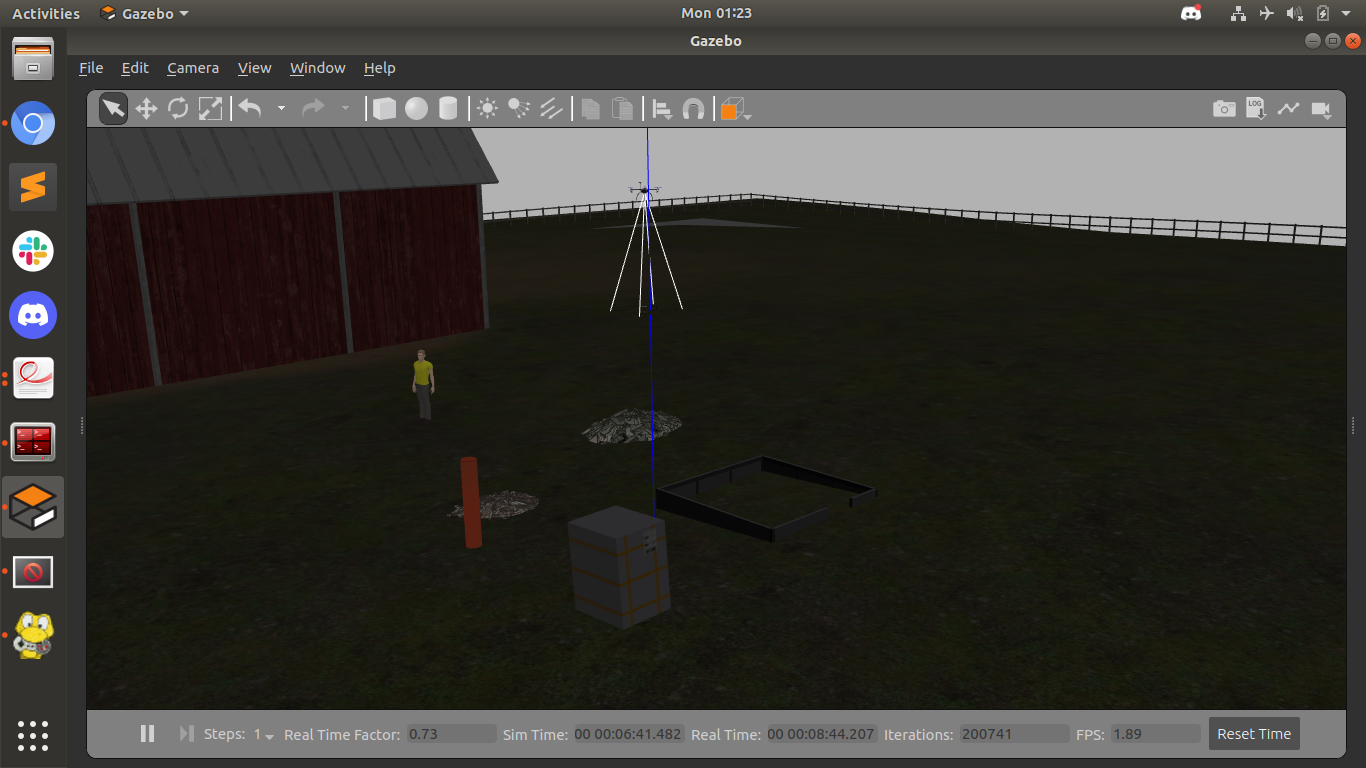}}
\caption{Simulated environment in Gazebo}
\label{gazebo}
\end{figure}

\begin{figure}[h]
\centerline{\includegraphics[width=0.45\textwidth]{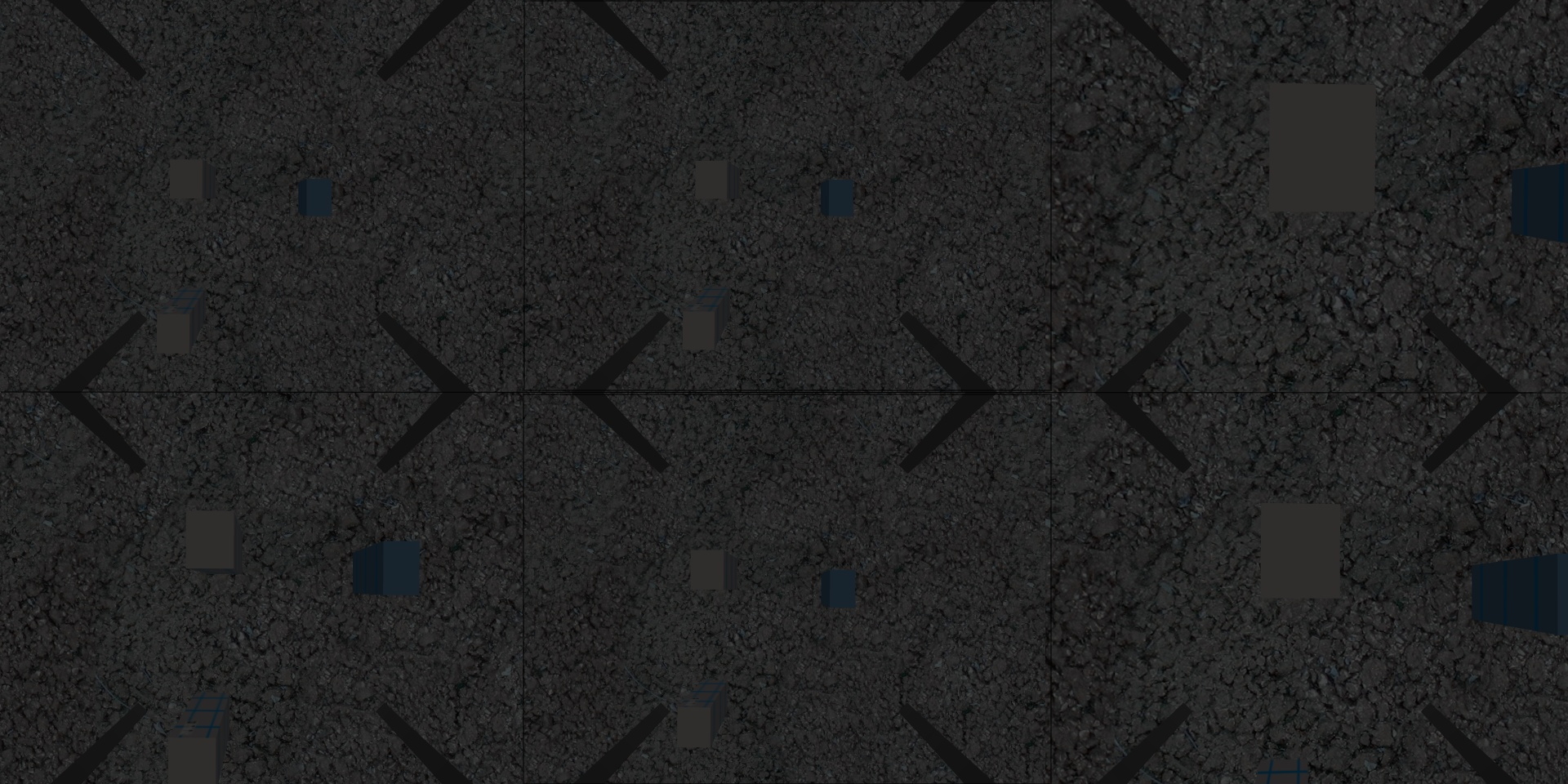}}
\caption{Representative images from the Dataset}
\label{dataset}
\end{figure}

We use the Gazebo simulator to perform evaluations of our approach. Gazebo enables us to perform realistic simulations using superior rendering and an accurate physics engine. Varied parameters such as illumination in the environment, landing sites, surface against which the landing site is simulated and the number of objects present in the vision cone are simulated. An example of a simulated environment is shown in Fig. \ref{gazebo} We use the PX4 framework\cite{px4} which is an open source flight control software. It is highly portable, Operating System independent and supports a variety of peripherals. Mavros, a MAVLink extendable communication node for ROS is used for communication with PX4. A DJI F450 frame, equipped with a GPS, a pixhawk 2.4.8 flight controller and a downward facing camera publishing images at 20 $Hz$ with a resolution of $640\times480$ is used.

\section{Evaluation Studies}\label{eval}
The algorithm is evaluated on the x86 architecture using a dual-core Intel Core i3-4010U CPU with a processor clocked at 1.7 GHz along with a 8GB DDR4 RAM on a laptop. These specifications are comparable to typical computers used on-board a MAV.
In the experiments, we evaluate the performance of our approach on five different categories of landing sites and scenarios,$L_n$, (Table \ref{tab_1}). The illumination scale varies from 1-10 according to environmental lighting used in Gazebo with 1 being extremely low lighting to 10 being extremely bright.

\begin{table}[h]
\caption{Landmarks}
\begin{center}
\begin{tabular}{|l|l|l|l|l|}
\hline
\textbf{Site} & \textbf{Illumination} & \textbf{Background} & \textbf{Length(m)} & \textbf{Breadth(m)} \\ \hline
\textbf{$L_1$}       & 2                     & Road                & 1.8                & 0.5                 \\ \hline
\textbf{$L_2$}       & 3                     & Road                & 0.7                & 0.8                 \\ \hline
\textbf{$L_3$}       & 4                     & Plain               & 1.2                & 1                   \\ \hline
\textbf{$L_4$}       & 7                     & Grass               & 0.5                & 0.5                 \\ \hline
\textbf{$L_5$}       & 9                     & Road                & 1.1                & 0.8                 \\ \hline
\end{tabular}
\label{tab_1}
\end{center}
\end{table}
Few images of the actual landing site are augmented to produce a dataset for the second stage of training, of size 25 images in our experiment. Using the weights available after our first stage of training, the model is further trained with the above images. The number of epochs for which the model is trained is shown in Table \ref{tab_2}. On our processor, these times varied from an average of $623s$ for 50 epochs to an average of $1837s$ for 100 epochs.
Our experiments investigate the robustness of our approach in different environmental illumination intensities and against different backgrounds while recording accuracy and computational usage parameters. We highlight an assumption made here that the accuracy is enough to ensure the target landing site comes into the field-of-view of the downward facing camera. Our results show that we achieve landing the MAV on certain significantly smaller landmarks in adverse environmental conditions. A summary of the results of has been shown in Table \ref{tab_2}. 

\begin{figure}[h]
\centerline{\includegraphics[width=0.45\textwidth]{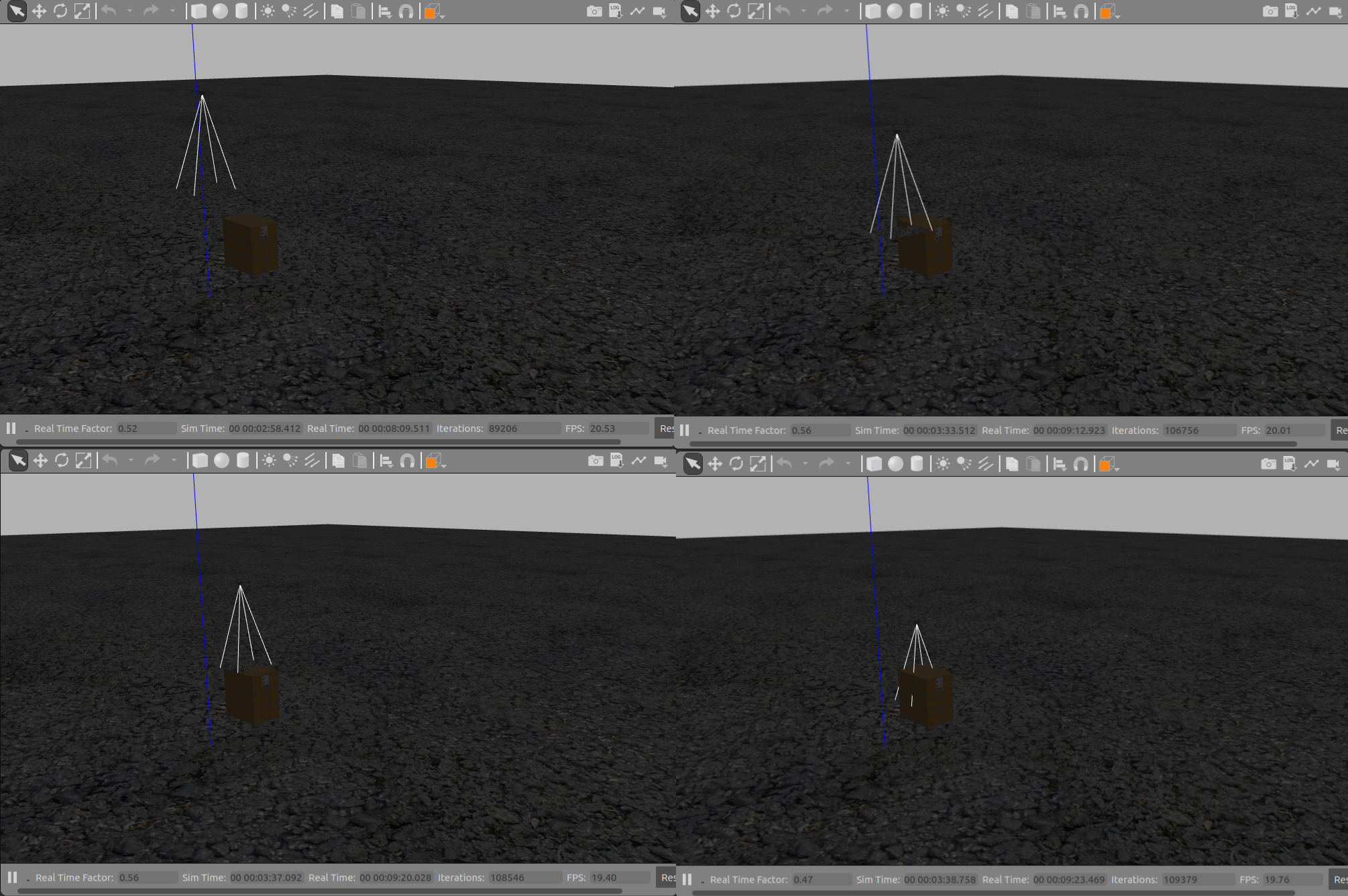}}
\caption{Sequence of frames showing Quadcopter landing}
\label{land}
\end{figure}

\begin{figure}[h]
\centerline{\includegraphics[width=0.45\textwidth]{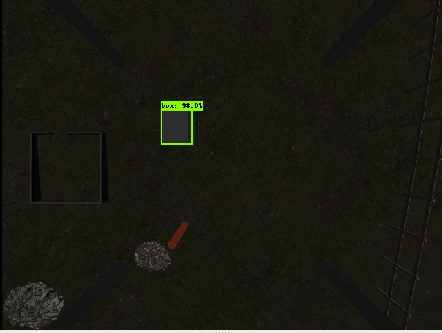}}
\caption{Site detection - Obstacles and Grass Background}
\label{site-detection}
\end{figure}

\begin{figure}[h]
\centerline{\includegraphics[width=0.42\textwidth]{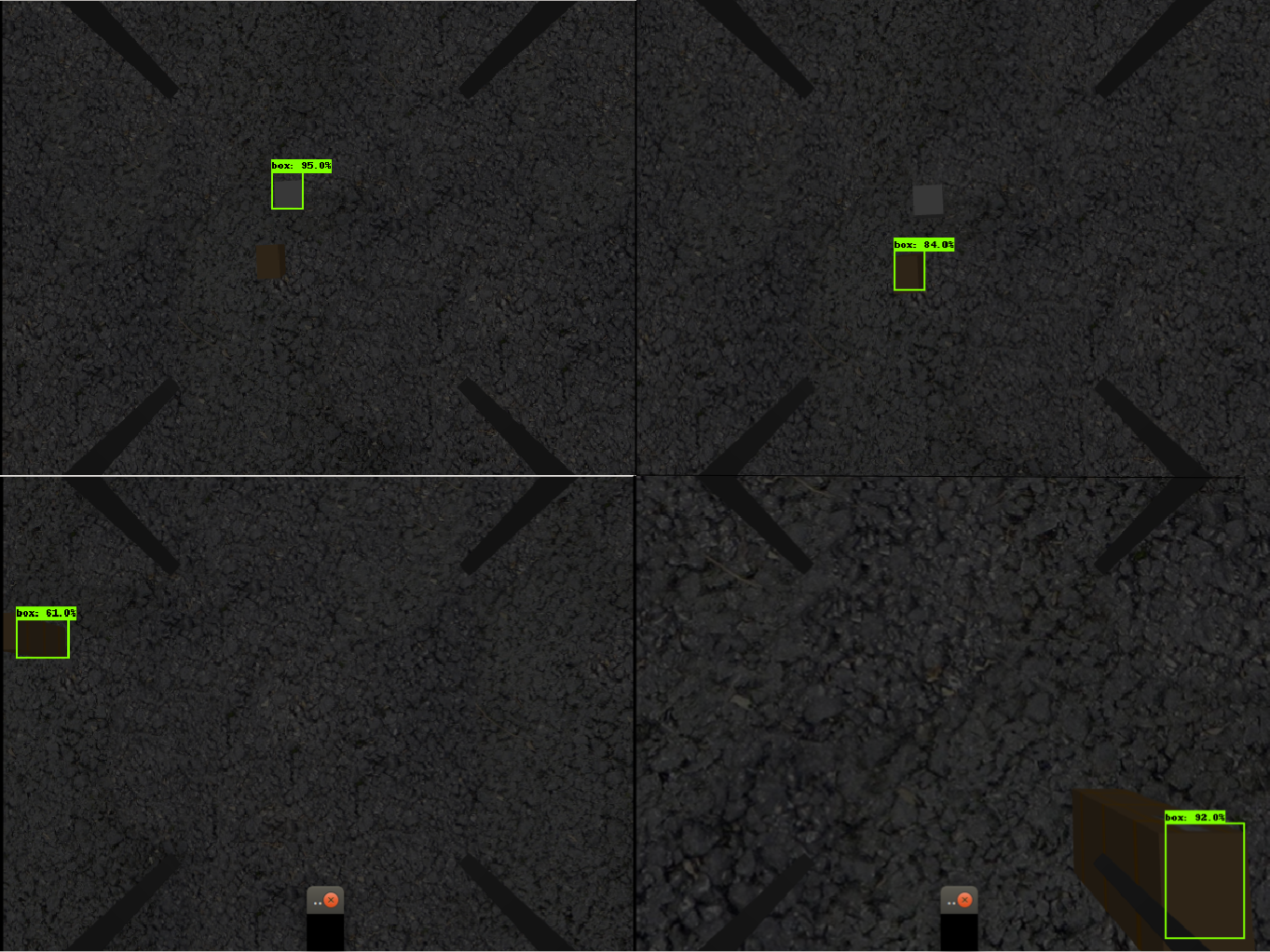}}
\caption{Corner Cases in Landing Site detection}
\label{corner}
\end{figure}

The average frame processing time increased in ideal illumination and was significantly lower in dimly lit environments. A similar trend was observed in CPU usage, while in terms of accuracy, the highest was obtained in scenario $L_4$ shown in Fig. \ref{site-detection} where illumination was adequate. The lowest accuracy was in $L_1$ proving among visual factors, illumination proves to be the most critical.
Apart from this even in cases of partial occlusion and in cases where an affine transformation existed due to the orientation of the MAV, the landing site was detected properly. In addition to scenario $L_4$ which had multiple visually different objects visible, we also test against multiple similar objects visible in the frame. These cases have been shown in Fig.\ref{corner}. As shown in the $top-left$ and $top-right$ image, the detector was trained for 50 epochs on two different landing sites and detects each of them accurately.  
A standard sequence of steps involved in landing has been shown in Fig.\ref{land}. 

\begin{table}[h]
\caption{Accuracy, Timing ,CPU Usage and Training Time}
\begin{center}
\begin{tabular}{|l|l|l|l|l|}
\hline
\textbf{Site} & \textbf{Accuracy (\%)} & \textbf{Avg. Time (ms)}& \textbf{CPU (\%)}& \textbf{Epochs} \\ \hline
\textbf{$L_1$}   & 92.1                   & 62.11   & 72.78 & 50   \\ \hline
\textbf{$L_2$}    & 97.5                   & 62.64  & 75.23 & 50  \\ \hline
\textbf{$L_3$}    & 96.8                   & 66.56  & 63.70 & 70  \\ \hline
\textbf{$L_4$}    & 98.4                   & 74.12  & 74.13 & 80  \\ \hline
\textbf{$L_5$}    & 95.2                   & 70.62  & 77.89 & 100  \\ \hline
\end{tabular}
\label{tab_2}
\end{center}
\end{table}
A comparison of our method with a GPS based method and fiducial tag based methods \cite{bc} and \cite {Araar} has been shown in table \ref{tab_3}. While the average accuracy is comparable, we out-perform the method in-terms of average landing time and landing speed.    
\begin{table}[h]
\caption{Comparative Table}
\begin{center}
\begin{tabular}{|l|l|l|l|}
\hline
\textbf{Source} & \textbf{Accuracy (m)} & \textbf{Landing Speed (m/s)} &\textbf{Time(s)} \\ \hline
Ours   & 0.16 & 0.25 & 41.7   \\ \hline
GPS-based  & 1-3 & 0.6 & - \\ \hline
Wubben et al. \cite{bc} & 0.11 & 0.1 & 55\\ \hline
Araar et al. \cite{Araar}& 0.13 & 0.06 & -\\ \hline
\end{tabular}
\label{tab_3}
\end{center}
\end{table}

\section{Conclusion}
In this work we have introduced a novel deep learning vision based MAV autonomous landing system augmented by a Kalman Filter. We implement a training strategy that allows passing the landing site prior to flight and our approach achieves high accuracy comparable to standard deep-learning based methods where the landing site is fixed. We implement a Kalman Filter that utilises the constraints imposed by the bounding box coordinates to achieve precise control for landing on the site. Our method works even on partially occluded sites and in a diverse set of environmental conditions. Future work includes predicting landing paths with geometrically tight trajectory constraints, implementation on moving platforms and adding the additional application of obstacle avoidance using the downward facing camera.

\vspace{12pt}
\end{document}